\documentclass{article}
\usepackage{ijcai26}

\usepackage{times}
\usepackage{url}
\usepackage[hidelinks]{hyperref}
\usepackage[utf8]{inputenc}
\usepackage[small]{caption}
\usepackage{graphicx}
\usepackage{amsmath}
\usepackage{amsthm}
\usepackage{booktabs}
\usepackage{multirow}

\title{Configurable Semantic Chunking for Biomedical Information Extraction in Retrieval-Augmented Generation}

\author{
Riya Ahuja$^{1,2,*}$
\and
Tim Kacprowski$^{1,2}$\and
Roya Shiasi Sardoabi$^{1,2,*}$\\
\affiliations
$^1$Institute of Data Science in Biomedicine, Technische Universit\"at Braunschweig, Germany\\
$^2$Braunschweig Integrated Centre of Systems Biology (BRICS), Technische Universit\"at Braunschweig, Germany\\
$^*$Corresponding authors: Riya Ahuja and Roya Shiasi Sardoabi\\
\emails
r.ahuja@tu-braunschweig.de,
t.kacprowski@tu-braunschweig.de,
roya.shiasi-sardoabi@tu-braunschweig.de
}

\begin{document}
\hyphenation{BioMedRAG BiomedRAG ChemProt MedLLaMA}
\maketitle

\begin{abstract}
BioMedRAG introduced retrieval-augmented generation with a learned chunk scorer for biomedical information extraction. However, it relies on fixed-size chunking which can fragment semantic evidence. We propose a configurable semantic chunking framework that addresses this limitation by combining entity-preserving windows, trigger-centered chunking, proposition-first extraction, tiered trigger prioritization, and hierarchical relation resolution. The framework integrates with BioMedRAG by replacing only the chunk construction stage while preserving the embedding model, learned chunk scorer, generator, and evaluation protocol. We evaluate the framework on biomedical relation extraction benchmarks (GM-CIHT, DDI, ChemProt) and adverse event classification (ADE). On GM-CIHT, the full hybrid configuration achieves 82.6\% F1, improving over the fixed-size baseline (74.2\% F1) by 8.4 points under our experimental setup. Cross-dataset analysis shows that semantic chunking improves extraction datasets with explicit relation cues, such as GM-CIHT and DDI, while fixed chunking remains competitive or stronger for dense biochemical extraction and binary classification settings such as ChemProt and ADE. By externalizing chunking logic into configuration files, the framework provides an interpretable and adaptable alternative to rigid fixed-size chunking for biomedical RAG pipelines.
\end{abstract}
\section{Introduction}
\label{sec:introduction}
Biomedical literature is expanding quickly, with PubMed now indexing more 
than 39 million articles~\cite{pubmed2026}. Extracting structured knowledge, 
such as drug-disease interactions, gene-protein relationships, and adverse drug 
events, from this vast corpus is critical for clinical decision support, drug 
discovery, and precision medicine~\cite{biomedical-nlp-survey2023}. Retrieval-augmented 
generation (RAG) has emerged as a powerful paradigm for this task~\cite{lewis2020rag}, 
combining large language models (LLMs) with external evidence retrieval to improve 
factual grounding and interpretability.

BioMedRAG~\cite{biomedrag2024} introduced a specialized RAG framework for biomedical information extraction and reports strong results on relation extraction benchmarks. However, it relies on fixed-size 
chunking, uniformly splitting sentences into 5-word windows regardless of semantic 
structure. This strategy frequently fragments relation-bearing expressions, resulting 
in incomplete or misaligned evidence. For example, the sentence \textit{``Aspirin treats headache by inhibiting prostaglandin synthesis''} may be split into chunks such as \textit{``Aspirin treats headache by inhibiting''} and \textit{``prostaglandin synthesis''}, separating the drug mechanism from its target. Such fragmentation reduces retrieval precision and forces the LLM to infer missing relational context.

To address this limitation, we propose a \textbf{configurable semantic chunking 
framework} that preserves meaningful semantic boundaries through an incremental design. 
Our framework builds incrementally on entity-aware segmentation that respects 
named entity boundaries. We progressively incorporate: (1)~relation-trigger-centered 
extraction with tiered prioritization to distinguish explicit triggers 
(\textit{treats}, \textit{inhibits}) from generic terms (\textit{medication}, 
\textit{drug}); (2)~hierarchical relation resolution to handle competing relation 
types; and (3)~proposition-first extraction that isolates minimal 
subject-trigger-object spans.
These components integrate through a unified scoring mechanism that balances trigger 
confidence, relation specificity, and contextual alignment, enabling seamless 
integration with BioMedRAG's trained chunk scorer.

Accurate extraction of biomedical relations from literature has direct 
implications for clinical practice and translational medicine. Drug-drug 
interaction detection (DDI) informs prescription safety and adverse event 
prevention, while chemical-protein binding relationships (ChemProt) 
accelerate drug target discovery and repurposing. Gene-disease associations 
extracted from biomedical texts support precision medicine and clinical 
genomics. By improving retrieval quality in RAG systems, our semantic 
chunking framework enhances the reliability of automated biomedical knowledge 
extraction, reducing manual curation burden and enabling real-time 
literature-based clinical decision support.

We evaluate our framework on four biomedical benchmarks from the BioMedRAG 
repository~\cite{biomedrag2024}: GM-CIHT, DDI, and ChemProt for triple extraction, 
and ADE for adverse event classification. Experimental results reveal substantial 
improvements on complex relational tasks requiring precise evidence localization, 
while detailed component-wise analysis characterizes the contribution of each 
framework element. Cross-dataset comparison provides practical insights into when 
semantic chunking benefits retrieval-augmented generation systems.

Our key contributions are summarized as follows:
\begin{enumerate}
    \item We propose a \textbf{configuration-driven semantic chunking framework} for biomedical retrieval-augmented information extraction. The framework replaces fixed-size windows with entity-preserving, trigger-aware, and proposition-oriented evidence candidates.

    \item We integrate the proposed chunking framework into BioMedRAG while keeping the embedding model, learned chunk scorer, generator, preprocessing pipeline, and evaluation protocol unchanged. This controlled design isolates the effect of chunk construction from other system components.

    \item We conduct an empirical evaluation on four biomedical benchmarks, including GM-CIHT, DDI, ChemProt, and ADE, and show that semantic chunking is most effective for extraction tasks with explicit relation cues and moderate entity density.

    \item We provide component-wise ablation studies, including the effects of entity-aware segmentation, proposition extraction, trigger prioritization, relation hierarchy resolution, ranking strategy, and NER-based entity detection.
\end{enumerate}

\section{Related Work}
\label{sec:related}

\subsection{Retrieval-Augmented Generation}

Large language models encode knowledge parametrically during training, but this knowledge is inherently limited: it becomes outdated, lacks domain-specific coverage, and can lead to hallucinated outputs~\cite{lewis2020rag}. Retrieval-augmented generation (RAG) mitigates these limitations by conditioning generation on externally retrieved evidence. Typical RAG pipelines operate in two stages: a retriever identifies relevant passages from a knowledge corpus, and a generator produces predictions conditioned on the retrieved evidence, improving factual accuracy and interpretability~\cite{lewis2020rag}.

Originally proposed for open-domain question answering~\cite{lewis2020rag}, RAG has since been applied to structured tasks such as knowledge base completion, dialogue systems, and code generation~\cite{shuster2021retrievaldialogue,zhou2023docprompting}. In biomedical NLP, accessing current literature is critical for accurate information extraction. BioMedRAG~\cite{biomedrag2024} applies this paradigm to biomedical relation extraction and is the system we extend; we describe its pipeline in Section~\ref{sec:preliminaries}.

\subsection{Text Chunking for Retrieval}

Text chunking strategies strongly influence retrieval quality in RAG 
systems~\cite{gao2023rag-survey}. Fixed-size chunking splits text into uniform 
windows, while sliding windows add overlap to preserve context. Recent work has 
investigated optimal retrieval granularity, comparing document, passage, sentence, 
and proposition-level units~\cite{chen2024dense}. However, these methods are 
designed for multi-sentence documents, making them ill-suited to biomedical 
relation extraction tasks, where relations are typically expressed within single 
sentences of 15-30 words.

Sub-sentence chunking methods aim to capture finer-grained semantics than 
sentence or paragraph-level segmentation. Proposition-based approaches decompose 
sentences into atomic, self-contained units for retrieval~\cite{chen2024dense,hosseini2024scalable}. 
Other work explores contextual chunk embeddings using long-context 
models~\cite{gunther2024latechunking}. These methods, however, are largely domain-agnostic and do not enforce biomedical entity preservation or explicit modeling of relation triggers.

\subsection{Biomedical Information Extraction}
Biomedical relation extraction aims to identify structured relationships 
between entities in scientific literature~\cite{zhou2014biomedrel}. 
Traditional approaches employ supervised learning with manually designed 
linguistic and semantic features~\cite{fundel2007relex} or neural 
architectures such as CNNs~\cite{zeng2014relationcnn}, 
LSTMs~\cite{zhou2016attentionlstm}, and Graph Neural 
Networks~\cite{zhu2019graphconvre}.
BioMedRAG~\cite{biomedrag2024} demonstrates that evidence quality 
critically influences extraction performance in retrieval-augmented 
biomedical systems.

\section{Method}
\label{sec:method}

\subsection{Preliminaries: The BioMedRAG Pipeline}
\label{sec:preliminaries}
We briefly review BioMedRAG~\cite{biomedrag2024}, the retrieval-augmented
extraction framework our approach extends.

BioMedRAG performs biomedical information extraction through
retrieval-augmented generation. For each input sentence $x$, it retrieves
the top-$k$ relevant evidence chunks from a corpus $\mathcal{D}$ and
generates triples conditioned on both $x$ and the retrieved
evidence~\cite{biomedrag2024}. Candidate chunks are stored in a relational
key--value memory, retrieved by embedding similarity, and re-ranked by a
chunk scorer trained to prefer evidence that improves downstream
prediction. Retrieval quality therefore depends fundamentally on chunk
quality.

BioMedRAG constructs its chunk database by splitting each sentence into
fixed 5-word windows with no overlap:
\begin{equation}
\label{eq:fixed}
\mathcal{C}_{\text{fixed}}(x) = \{(w_i, \dots, w_{i+4}) \mid i = 1, 6, 11, \dots\},
\end{equation}
which ignores semantic structure. This strategy frequently fragments
entities, for instance, the hyphenated compound \emph{alpha-ketoglutarate}
may be split into \emph{alpha-} and \emph{ketoglutarate} across separate
5-word windows, and breaks relation-bearing propositions across chunk
boundaries, resulting in incomplete or misaligned evidence for the language
model. Our framework replaces this chunk-construction stage; the embedding
model, learned chunk scorer, generator, and evaluation protocol remain
unchanged.

\subsection{Problem Formulation}

We address two biomedical information extraction tasks. The primary task is 
\textbf{triple extraction}: given a sentence $x = (w_1, \ldots, w_n)$, extract 
structured relations $(h, r, t)$ where $h$ is the subject entity (head), 
$r \in \mathcal{R}$ is a relation type, and $t$ is the object entity (tail). 
For example, from \textit{``aspirin inhibits prostaglandin synthesis''}, we 
extract $\langle$\textsc{aspirin}, \textsc{inhibits}, 
\textsc{prostaglandin}$\rangle$. The secondary task is \textbf{relation 
classification}: given a sentence with marked entity spans, predict the 
relation type between them. Our chunking framework addresses both 
tasks, as both require identifying relation-bearing evidence within sentences.

\subsection{Configurable Semantic Chunking Framework}
\label{sec:config-framework}

Our framework mitigates the fragmentation introduced by fixed-width chunking by
constructing a hybrid candidate pool and selecting evidence with an explicit
bias toward structurally complete relation spans.

\paragraph{Stage 1: Multi-Source Candidate Generation.}
For each sentence, we generate candidate chunks from three sources.
(i) \emph{Entity-aware sliding windows} preserve biomedical entity boundaries and
capture relation-trigger context (Sections~\ref{sec:entity-aware} and
\ref{sec:relation-triggers}).
(ii) \emph{Proposition-first extraction} isolates minimal subject--trigger--object
spans (Section~\ref{sec:proposition}); overlapping propositions are internally
prioritized using tiered trigger specificity and relation hierarchy
(Sections~\ref{sec:tiered-triggers} and \ref{sec:hierarchy}).
(iii) \emph{Fixed-width fallback windows} (the original 5-word splits) are added 
to ensure non-empty coverage when entity/trigger signals are weak.

\paragraph{Stage 2: Similarity-Based Selection with Proposition Bias.}
Candidates are ranked by cosine similarity between chunk embeddings and the
target relation definition, consistent with the embedding-based retrieval step
in BioMedRAG~\cite{biomedrag2024}. We then apply a \emph{proposition bias}: if at
least one proposition-based candidate appears among the top-$N$ ranked candidates,
we promote the best such proposition into the final top-$k$ set. This guarantees
that the downstream generator receives at least one chunk with verified
subject--trigger--object structure, while retaining high-similarity contextual
chunks for coverage. The selected chunks are finally passed to BioMedRAG's
trained chunk scorer for re-ranking.

\paragraph{Configurability.}
All dataset-specific knowledge, including trigger vocabularies, tier definitions,
relation hierarchies, negation rules, and context markers, is externalized in JSON
configuration files. This design makes the chunking decisions explicit and
inspectable, enabling adaptation to new biomedical relation sets without code
changes.

\paragraph{Shared Trigger Lexicon.} Both entity-aware windows and proposition 
extraction use a common tiered trigger vocabulary for relation signal detection, 
ensuring consistent identification of relation-bearing spans across candidate types.

\subsection{Entity-Aware Segmentation}
\label{sec:entity-aware}
This subsection describes the first candidate source in Stage~1: entity-aware
sliding windows.

Biomedical named entities (drugs, proteins, diseases) frequently span multiple
tokens and often include hyphenation (e.g., \textit{alpha-ketoglutarate},
\textit{acetyl-CoA carboxylase}). Fixed-width chunking can cut through such
entities, producing partial strings that degrade retrieval similarity and
downstream relation evidence.

\paragraph{Entity Detection.}
We detect entity spans using lightweight pattern rules: bracketed markup (e.g.,
\textit{[aspirin]}), capitalized multi-token terms (e.g., \textit{Tumor Necrosis
Factor}), and hyphenated compounds (e.g., \textit{alpha-ketoglutarate}). Each
entity $e$ is represented as a character span $(e_{\text{start}}, e_{\text{end}})$.

\paragraph{Scope of Entity Detection.}
This step is intended as a boundary-preservation mechanism rather than as a complete biomedical named entity recognition module. The detected spans are used only to adjust chunk boundaries and construct proposition candidates; final relation prediction is still performed by the BioMedRAG scorer and generator. This design favors low overhead and direct integration with the existing pipeline, but it may miss lowercase or irregular biomedical mentions. To examine this limitation, we compare pattern-based entity spans with NER-based spans in the ablation study.

\paragraph{Sliding Window Adjustment.}
Given a window size $w$ and stride $s$, we adjust window boundaries whenever a
boundary falls inside an entity span: (i) if the window end intersects an entity,
we extend it to include the full entity; (ii) if the window start intersects an
entity, we shift the start past the entity. This guarantees that no entity is
split across chunks.

\paragraph{Example.}
Consider \textit{``Tumor necrosis factor alpha receptor-complex regulates immune
responses.''} With fixed 5-token windows, one chunk may end at
\textit{``... alpha receptor-''} while the next begins with \textit{``complex ...''},
splitting the hyphenated entity \textit{receptor-complex}. Our entity-aware approach moves the boundary so that \textit{tumor necrosis factor alpha receptor-complex} remains intact within a single chunk.

\subsection{Relation Trigger Detection}
\label{sec:relation-triggers}

Entity-aware segmentation preserves entity boundaries but remains agnostic to
whether a chunk actually expresses a semantic relation. We therefore augment the
sliding window strategy with trigger-based detection to emphasize relation-bearing
content.

\paragraph{Trigger Lexicon Construction.}
For each dataset, candidate triggers are collected from the training split by identifying words and short phrases that occur between or near annotated entity pairs. These candidates are grouped by their associated gold relation labels and inspected for relation specificity. Terms that frequently co-occur with a single relation are retained as stronger triggers, whereas terms appearing across several relation types are assigned to lower tiers or removed when they provide limited discriminative value. Morphological variants expressing the same cue are merged, for example, ``inhibit'', ``inhibits'', ``inhibited'', and ``inhibition''. For instance, the \textsc{Inhibits} relation may include triggers such as ``inhibits'', ``blocks'', ``suppresses'', ``antagonizes'', ``downregulates'', and ``attenuates'', depending on the dataset-specific relation definition. If no trigger is detected, the framework falls back to entity-aware sliding windows and fixed-width chunks to preserve coverage.

\paragraph{Trigger-Centered Chunking.}
When a trigger is detected, we generate a local context window of $\pm4$ words
around it to capture nearby entities and modifiers. For example, in
\textit{``Aspirin effectively inhibits prostaglandin synthesis in tissues,''}
the trigger \textit{inhibits} produces a chunk that captures both the agent
(\textit{Aspirin}) and the target (\textit{prostaglandin synthesis}). The window
size is intentionally small to emphasize local relational evidence.

\paragraph{Limitation.}
This approach treats all triggers equally, regardless of semantic specificity.
Consequently, a chunk containing a generic term such as \textit{affects} receives
the same priority as one containing a precise term such as \textit{inhibits}. This
motivates tiered trigger prioritization, introduced next
(Section~\ref{sec:tiered-triggers}).
\subsection{Tiered Trigger Prioritization}
\label{sec:tiered-triggers}

Not all relation triggers carry equal semantic weight. Generic terms such as
\textit{affects} or \textit{modulates} are ambiguous, while specific terms like
\textit{inhibits} or \textit{therapy} convey precise mechanistic or clinical
meaning. We therefore organize triggers into three tiers based on semantic
specificity and directional clarity, following common distinctions in biomedical
relation extraction between explicit actions, contextualized actions, and generic
associations.

\paragraph{Trigger Tier Definitions.}
For each relation type, we define:
\begin{itemize}
    \item \textbf{Primary triggers} (tier bonus 3): Highly specific terms that 
    unambiguously indicate the relation (e.g., \textit{inhibits}, \textit{therapy},
    \textit{activates}).
    \item \textbf{Secondary triggers} (tier bonus 2): Moderately specific synonyms 
    or related terms that express the relation with reduced precision (e.g., 
    \textit{blocks}, \textit{prescribed}, \textit{enhances}).
    \item \textbf{Tertiary triggers} (tier bonus 1): Generic or weaker signals 
    that may indicate a relation but lack sufficient specificity when used in 
    isolation (e.g., \textit{reduces}, \textit{medication}, \textit{associated}).
\end{itemize}

\paragraph{Example.}
For \textsc{treats}, primary triggers include \textit{treats},
\textit{treatment}, and \textit{therapy}; secondary triggers include
\textit{prescribed}, \textit{administered}, and \textit{therapeutic}; and tertiary
triggers include \textit{medication} and \textit{drug}. The sentence
\textit{``Aspirin therapy alleviates headache''} contains a primary trigger
(\textit{therapy}), while \textit{``Aspirin medication alleviates headache''}
contains only a tertiary trigger (\textit{medication}).

\paragraph{Usage.}
Tier membership is used during proposition-first extraction
(Section~\ref{sec:proposition}) to rank overlapping candidate propositions. When
multiple propositions cover the same text span, the proposition containing a
higher-tier trigger is retained, ensuring that the most semantically precise
relational evidence is passed to the selection stage.

\paragraph{Tier Assignment.}
Triggers were assigned to tiers based on two criteria: (i)~corpus frequency 
analysis of trigger--relation co-occurrence in training data, where high 
co-occurrence with a single relation indicates specificity, and 
(ii)~linguistic directness, where verb forms (e.g., \textit{inhibits}) are 
preferred over nominalizations (e.g., \textit{inhibition}) and generic 
associations (e.g., \textit{affects}). Tier assignments were validated on the 
development set (Table~\ref{tab:datasets}) by comparing proposition extraction precision across alternative 
classifications.

\subsection{Relation Hierarchy Resolution}
\label{sec:hierarchy}

Building on tiered trigger prioritization, we address a second source of ambiguity:
sentences that express multiple biomedical relations simultaneously. For example,
\textit{``aspirin treats inflammation by inhibiting COX-2''} contains both a
\textsc{treats} relation (aspirin--inflammation) and an \textsc{inhibits} relation
(aspirin--COX-2). Without explicit disambiguation, extracted propositions may
emphasize relations that are less informative for the target extraction objective.

We resolve this ambiguity through configurable relation hierarchies that encode
dataset- and task-specific priorities. For GM-CIHT, we define the hierarchy shown
in Table~\ref{tab:hierarchy}.

\begin{table}[t]
\centering
\small
\begin{tabular}{lc}
\toprule
\textbf{Relation} & \textbf{Priority weight} \\
\midrule
\textsc{treats} & 4 \\
\textsc{inhibits}, \textsc{stimulates}, \textsc{causes}, \textsc{prevents} & 3 \\
\textsc{affects}, \textsc{interacts\_with}, \textsc{reduces}& 2 \\
\textsc{coexists\_with} & 1 \\
\bottomrule
\end{tabular}
\caption{Relation hierarchy for GM-CIHT. Higher weights indicate higher
precedence during proposition ranking. Weights were tuned on the
development set; alternative orderings yielded lower F1.}
\label{tab:hierarchy}
\end{table}
\paragraph{Priority Assignment.}
Relation weights reflect two explicit criteria:
(i)~\textit{semantic specificity}, indicating how directly a relation encodes an
actionable interaction between entities, and
(ii)~\textit{task relevance}, reflecting the importance of the relation for the
target extraction objective. In GM-CIHT, \textsc{treats} receives the highest
weight as the dataset emphasizes therapeutic relationships. Directional
mechanistic relations such as \textsc{inhibits} and \textsc{stimulates} are
informative but secondary, while \textsc{coexists\_with} denotes weak associative
evidence. 

\paragraph{Hierarchy Tuning.}
Relation weights were determined through grid search on the development
set (Table~\ref{tab:datasets}): we evaluated all permutations of relation
orderings and selected the configuration maximizing extraction F1. The
reported hierarchy in Table~\ref{tab:hierarchy} outperformed flat
(equal-weight) and inverted orderings by 1.2--2.1\% F1. For other datasets,
relation weights are specified through dataset-specific configuration files
using the same principle, while the core chunking algorithm and selection
strategy remain unchanged.

\paragraph{Example.}
Consider \textit{``Metformin treatment reduces glucose levels by activating the
AMPK pathway in diabetic patients.''} This sentence supports multiple relations:
\begin{itemize}
    \item \textsc{treats}: \textit{treatment} (primary trigger), clinical context
    (\textit{diabetic patients})
    \item \textsc{stimulates}: \textit{activating} (primary trigger)
    \item \textsc{affects}: \textit{reduces} (secondary trigger)
\end{itemize}
Although all relations are supported by valid triggers, the hierarchy ensures that
the proposition emphasizing the therapeutic relation (metformin--diabetes) is
retained over mechanistic or associative alternatives.

\paragraph{Combined Disambiguation.}
Relation hierarchy resolution operates jointly with tiered trigger prioritization
during proposition-first extraction (Section~\ref{sec:proposition}). When multiple
propositions are extracted from the same sentence, they are ranked
lexicographically: propositions associated with higher-priority relations are
preferred, and ties are resolved by trigger tier strength. Only non-overlapping
propositions with the highest precedence are retained.

\subsection{Proposition-First Extraction}
\label{sec:proposition}

Sliding windows may capture entity context or relation context separately, 
but fail to preserve complete relational evidence in a single chunk. For 
example, \textit{``Metformin, a widely used antidiabetic, improves insulin 
sensitivity in diabetic patients''} may yield windows containing the entity 
(\textit{``Metformin...antidiabetic''}) or the relation (\textit{``improves 
insulin sensitivity''}), but neither forms a complete triple.

We address this through \textbf{proposition-first extraction}, inspired 
by~\cite{hosseini2024scalable}. We extract candidate chunks corresponding 
to atomic propositions: self-contained units containing a subject, a relation 
trigger, and an object.

\paragraph{Proposition Patterns.} We detect propositions using three syntactic 
patterns:
\begin{itemize}
    \item \textbf{Infix}: Entity$_1$ -- \textsc{trigger} -- Entity$_2$ 
    (e.g., \textit{``Aspirin inhibits prostaglandins''});
    \item \textbf{Prefix}: \textsc{trigger} -- Entity$_1$ \ldots Entity$_2$ 
    (e.g., \textit{``Treatment of diabetes with metformin''});
    \item \textbf{Postfix}: Entity$_1$ \ldots Entity$_2$ -- \textsc{trigger} 
    (e.g., \textit{``Aspirin--COX-2 interaction''}).
\end{itemize}
For each trigger, we identify nearest entities within a bounded context 
($\pm 25$ tokens) and extract the minimal span covering subject, trigger, 
and object. We expand each span by $\pm 3$ tokens to capture adjacent 
negation markers (e.g., \textit{does not}, \textit{fails to}) and uncertainty 
hedges (e.g., \textit{may}, \textit{possibly}) that can invert or weaken the 
expressed relation, ensuring the downstream model receives complete polarity 
information.

\paragraph{Pattern Coverage.} These patterns cover the majority of explicit 
relation expressions in biomedical corpora. Complex constructions such as 
nested relations or coordinated arguments are handled implicitly: sliding 
window candidates provide fallback coverage when proposition patterns do 
not match.

\paragraph{Example.} Consider the sentence: \textit{``Aspirin therapy reduces 
inflammation by inhibiting COX-2 expression.''} Two propositions are extracted:
\begin{enumerate}
    \item \textit{``Aspirin therapy reduces inflammation''} - Trigger: 
    \textit{reduces} (tertiary for \textsc{affects}), Priority: $2 + 1 = 3$
    \item \textit{``Aspirin inhibiting COX-2''} - Trigger: \textit{inhibiting} 
    (primary for \textsc{inhibits}), Priority: $3 + 3 = 6$
\end{enumerate}
The second proposition is ranked higher due to its stronger trigger and 
higher-priority relation, ensuring precise mechanistic evidence is retained.

\paragraph{Passive Handling.} Passive constructions (e.g., \textit{``treated 
by''}) reverse surface entity order. We detect passive triggers followed by 
markers (\textit{by}, \textit{with}) and invert semantic roles accordingly.

\paragraph{Ranking and Integration.} Overlapping propositions are deduplicated 
by retaining those with the highest combined priority:
\begin{equation}
    P(p) = \text{Weight}_{\text{hierarchy}}(r) + \text{Bonus}_{\text{tier}}(t)
    \label{eq:priority}
\end{equation}
where $r$ is the detected relation type, $t$ is the trigger word, 
$\text{Weight}_{\text{hierarchy}}(r) \in \{1, 2, 3, 4\}$ follows 
Table~\ref{tab:hierarchy}, and $\text{Bonus}_{\text{tier}}(t) \in \{3, 2, 1\}$ 
for primary, secondary, and tertiary triggers respectively. This formula 
integrates linguistic confidence (trigger specificity) with clinical importance 
(relation hierarchy) into a unified ranking score.

Proposition chunks complement sliding windows: propositions ensure structural 
completeness while windows preserve broader context. Both candidate types are 
passed to the hybrid selection stage (Section~\ref{sec:hybrid-selection}).

\subsection{Hybrid Selection Strategy}
\label{sec:hybrid-selection}

The previous stages produce a diverse pool of candidates: entity-aware sliding 
windows, proposition spans (ranked using tiered triggers and relation hierarchy), 
and fixed-width fallback windows. We employ a hybrid strategy to select the 
final top-$k$ chunks, integrating semantic scoring with structural prioritization.

\paragraph{Scoring.} Following BioMedRAG~\cite{biomedrag2024}, all candidates are 
scored using cosine similarity between their averaged token embeddings and the 
target relation definition embedding.

\paragraph{Structural Prioritization via Proposition Bias.} Embedding similarity 
alone may favor verbose chunks over precise propositions. To ensure that the 
contributions of \textbf{tiered triggers} and \textbf{relation hierarchy} 
(Sections~\ref{sec:tiered-triggers}--\ref{sec:hierarchy}) propagate to the final 
selection, we apply a \textit{proposition bias}:
\begin{enumerate}
    \item \textbf{Slot~1 (Semantic Best):} The highest-similarity candidate is 
    selected for broad context.
    \item \textbf{Slot~2 (Structural Best):} If a proposition candidate, already 
    ranked by combined priority $P(p)$ (Equation~\ref{eq:priority}), exists in 
    the top-$N$ ($N{=}5$), it is promoted to this slot. This ensures that chunks 
    containing high-tier triggers and high-priority relations are retained 
    even if their embedding score is not the highest.
    \item \textbf{Fallback:} Otherwise, the second-highest similarity chunk 
    is selected.
\end{enumerate}

\paragraph{Design Rationale.}
Tiered triggers and relation hierarchy are applied during proposition extraction 
rather than entity-aware window generation. This reflects their distinct roles: 
entity-aware windows provide \textit{broad contextual coverage}, capturing 
surrounding entities and modifiers that inform the language model even without 
explicit relation signals. Propositions, in contrast, target \textit{precise 
relational evidence} where multiple competing triggers frequently co-occur 
within minimal spans. Tiered prioritization and hierarchy resolution are most 
effective when disambiguating such conflicts, which are common in propositions but rare 
in broader windows. Furthermore, applying internal scoring to entity-aware 
windows risks over-filtering useful contextual chunks that lack explicit 
triggers. The proposition bias ensures structurally complete evidence (ranked 
by tiers and hierarchy) reaches final selection, while entity-aware windows 
contribute complementary context ranked by semantic similarity alone.

\paragraph{Integration.} Selected chunks are passed to BioMedRAG's trained 
chunk scorer for final re-ranking~\cite{biomedrag2024}, maintaining full 
pipeline compatibility.

\section{Experiments}
\label{sec:experiments}

\subsection{Datasets}
\label{sec:datasets}

We evaluate on four biomedical datasets (Table~\ref{tab:datasets}): GM-CIHT, DDI, ChemProt (triple extraction), and ADE (binary relation classification). As in the BioMedRAG evaluation setting, we do not consider link prediction, where inputs are too short to admit meaningful sub-sentence chunking.

\begin{table}[t]
\centering
\small
\begin{tabular}{lccccc}
\toprule
\textbf{Dataset} & \textbf{Task} & \textbf{Rels} & \textbf{Train} & \textbf{Dev} & \textbf{Test} \\
\midrule
GM-CIHT & Extraction & 22 & 3,734 & 492 & 465 \\
DDI & Extraction & 4 & 1,027 & 258 & 1,094 \\
ChemProt & Extraction & 5 & 4,111 & 2,411 & 3,438 \\
ADE & Classification & 2 & 4,000 & 975 & 497 \\
\bottomrule
\end{tabular}
\caption{Dataset statistics. \textbf{Rels} denotes the number of relation types;
\textbf{Train}, \textbf{Dev}, and \textbf{Test} denote the number of instances
(sentence-label pairs) in each split. GM-CIHT covers 22 general biomedical
relations (therapeutic, mechanistic, associative). DDI focuses on drug-drug
interactions. ChemProt targets chemical-protein bindings with subtle semantic
distinctions. ADE is binary adverse event detection.}
\label{tab:datasets}
\end{table}

GM-CIHT serves as the primary development dataset for designing the general semantic chunking strategy, including entity-aware windows, trigger-tier usage, relation hierarchy resolution, and proposition bias. For each benchmark, dataset-specific symbolic resources such as relation labels, trigger vocabularies, tier assignments, and hierarchy definitions are specified through external configuration files. We do not otherwise tune the chunk scorer or generator on DDI, ChemProt, or ADE. All corpora are converted to a unified JSONL schema (\texttt{SENTENCE}, \texttt{SUBJECT\_TEXT}, \texttt{OBJECT\_TEXT}, \texttt{PREDICATE}; ADE is mapped from its native format) so chunking, embedding, and generation share a consistent interface.

\subsection{Baselines and Experimental Setup}
\label{sec:setup}

\paragraph{Baseline.} We compare against \textbf{BioMedRAG (Fixed)}~\cite{biomedrag2024}, which uses fixed 5-word chunking. This baseline is selected to isolate the effect of chunk construction while keeping the remaining pipeline unchanged, including the embedding model, learned chunk scorer, generator, preprocessing, and evaluation protocol. Thus, the comparison evaluates whether semantically structured evidence units improve BioMedRAG-style retrieval over fixed-size windows. Comparisons with sentence-level chunking, dependency-based span extraction, and external proposition segmentation methods are left for future work.

\paragraph{Implementation.} We use MedLLaMA-13B for computing chunk and relation embeddings with 512-token maximum length and mean-pooling. Following BioMedRAG~\cite{biomedrag2024}, we train chunk scorers initialized from Llama-2-13B with LoRA~\cite{hu2022lora} ($r{=}8$, $\alpha{=}8$, dropout $0.1$) for 1,000 steps using AdamW with bfloat16 precision. Separate scorers are trained per dataset.

\paragraph{Chunking Parameters.} Entity-aware windows use size $w{=}8$, stride $s{=}3$, bounds $[5, 12]$ tokens. Proposition extraction searches $\pm25$ tokens around triggers with $\pm3$ token context margins. Proposition bias threshold $N{=}5$. These parameters are kept fixed across datasets to support controlled comparison; systematic sensitivity analysis of window size, context range, and proposition bias threshold is left for future work.

\paragraph{Configuration.}
The core chunking algorithm, window parameters, proposition bias threshold, embedding model, learned scorer, generator, preprocessing pipeline, and evaluation protocol are kept fixed across datasets. Dataset-specific symbolic resources are provided through external JSON configuration files.

\paragraph{Evaluation.}
Following BioMedRAG~\cite{biomedrag2024}, we report micro-averaged Precision, Recall, and F1 for the extraction tasks GM-CIHT, DDI, and ChemProt. For these tasks, a prediction is considered correct only if the head span, relation type, and tail span exactly match the ground truth. For ADE, we report binary classification performance using the provided entity spans. All experiments use NVIDIA A40 GPUs with 4-bit NF4 quantization and fixed random seeds.

\subsection{Main Results}
\label{sec:main-results}

Table~\ref{tab:main-results} presents results across all four datasets.

\begin{table}[t]
\centering
\small
\begin{tabular}{llccc}
\toprule
\textbf{Dataset} & \textbf{Method} & \textbf{P} & \textbf{R} & \textbf{F1} \\
\midrule
\multirow{2}{*}{GM-CIHT} 
  & Fixed & 74.6 & 73.8 & 74.2 \\
  & \textbf{Ours} & \textbf{82.6} & \textbf{82.6} & \textbf{82.6} \\
\midrule
\multirow{2}{*}{DDI} 
  & Fixed & 78.2 & 78.2 & 78.2 \\
  & \textbf{Ours} & \textbf{79.2} & \textbf{79.2} & \textbf{79.2} \\
\midrule
\multirow{2}{*}{ChemProt} 
  & \textbf{Fixed} & \textbf{87.7} & \textbf{87.0} & \textbf{87.4} \\
  & Ours & 87.0 & 86.3 & 86.6 \\
\midrule
\multirow{2}{*}{ADE} 
  & \textbf{Fixed} & \textbf{87.0} & \textbf{90.7} & \textbf{88.8} \\
  & Ours & 86.1 & 88.6 & 87.3 \\
\bottomrule
\end{tabular}
\caption{Main results (\%). \textbf{P}, \textbf{R}, and \textbf{F1} denote
micro-averaged Precision, Recall, and F1-score. \textbf{Fixed} is the BioMedRAG
fixed 5-word chunking baseline; \textbf{Ours} is the proposed semantic chunking
framework. Our semantic chunking achieves gains on GM-CIHT (+8.4 F1) and DDI
(+1.0 F1), while fixed chunking remains competitive on ChemProt and ADE.}
\label{tab:main-results}
\end{table}

Our semantic chunking substantially improves extraction performance on GM-CIHT (\textbf{+8.4} F1 points: 74.2 $\rightarrow$ 82.6) and DDI (\textbf{+1.0} F1: 78.2 $\rightarrow$ 79.2). These datasets contain diverse relation types where tiered trigger prioritization and proposition extraction effectively disambiguate competing interpretations. On GM-CIHT, gains in both precision and recall indicate improved evidence quality without sacrificing coverage.

On ChemProt and ADE, fixed chunking remains competitive, outperforming 
our method by $0.8\%$ and $1.5\%$ F1 respectively. ChemProt involves 
fine-grained mechanistic distinctions (e.g., \textsc{agonist} vs.\ 
\textsc{activator}) that benefit from broader contextual cues preserved 
by fixed windows, while ADE is a binary classification task where 
comprehensive sentence-level context is more informative than targeted 
propositions. These results highlight that semantic chunking is most 
beneficial in settings with relational ambiguity rather than uniformly 
dense supervision, a pattern we analyze further in 
Section~\ref{sec:analysis}.

\subsection{Ablation Studies}
\label{sec:ablation}

We analyze component contributions and in-context learning sensitivity on GM-CIHT.

\paragraph{Component Ablation.}
Table~\ref{tab:ablation-components} shows incremental gains from each component.

\begin{table}[t]
\centering
\small
\begin{tabular}{lcc}
\toprule
\textbf{Configuration} & \textbf{F1} & \textbf{$\Delta$} \\
\midrule
Fixed (Baseline) & 74.2 & --- \\
+ Entity-Aware & 76.1 & +1.9 \\
+ Proposition & 76.3 & +0.2 \\
+ Tiered Triggers & 78.7 & +2.4 \\
+ Hierarchy (Full) & \textbf{82.6} & +3.9 \\
\bottomrule
\end{tabular}
\caption{Incremental component ablation on GM-CIHT (\%). Each row adds one
component to the configuration above it. \textbf{F1} is the micro-averaged
F1-score; \textbf{$\Delta$} is the change relative to the preceding row.}
\label{tab:ablation-components}
\end{table}

Entity-aware chunking provides a strong foundation (+1.9 F1) by preventing fragmentation. Proposition extraction adds a small gain (+0.2 F1), and tiered triggers yield a larger step (+2.4 F1). The relation hierarchy yields the largest incremental gain (+3.9 F1) by aligning evidence
selection with GM-CIHT’s therapeutic focus, consistently prioritizing
\textsc{treats} relations over weaker associations such as
\textsc{coexists\_with} when multiple interpretations are present.

\paragraph{In-Context Learning Sensitivity.}
Table~\ref{tab:ablation-icl} examines prompting requirements.

\begin{table}[t]
\centering
\small
\begin{tabular}{lccc}
\toprule
\textbf{Method} & \textbf{1 Ex.} & \textbf{3 Ex.} & \textbf{$\Delta$} \\
\midrule
Fixed & 74.2 & 75.1 & +0.9 \\
Ours & 82.6 & 77.2 & $-$5.4 \\
\bottomrule
\end{tabular}
\caption{Effect of the number of in-context examples on GM-CIHT (\%).
\textbf{1 Ex.} and \textbf{3 Ex.} denote micro-averaged F1 with one and three
in-context demonstrations, respectively; \textbf{$\Delta$} is the change from
one to three.}
\label{tab:ablation-icl}
\end{table}

Fixed chunking improves with additional examples (+0.9 F1), while our method performs best with one example, degrading with three ($-$5.4 F1). This suggests semantically coherent chunks saturate the model's contextual needs with fewer demonstrations, potentially reducing inference costs compared to lower-quality evidence requiring more examples; the drop with three examples may also reflect context limits or sensitivity to demonstration choice.

\paragraph{Ranking variant.}
Besides cosine similarity with proposition bias (Section~\ref{sec:hybrid-selection}), we evaluate a ranking that mixes $70\%$ cosine similarity with $30\%$ of a structural score derived from tiered triggers and relation hierarchy. Table~\ref{tab:ablation-ranking} shows that this mixed ranking reduces GM-CIHT F1 to $79.4\%$ ($-3.2$ vs.\ cosine with proposition bias at $82.6\%$). Proposition bias therefore appears sufficient to leverage structural signals: promoting a proposition into the final top-$k$ when it appears among the top candidates by cosine avoids diluting or mis-scaling the semantic similarity signal; the $70/30$ weighting may also be suboptimal or misaligned between score scales.

\begin{table}[t]
\centering
\small
\begin{tabular}{lcc}
\toprule
\textbf{Ranking} & \textbf{F1} & \textbf{$\Delta$} \\
\midrule
Cosine + proposition bias & 82.6 & --- \\
$70\%$ cosine + $30\%$ tiered/hierarchy & 79.4 & $-$3.2 \\
\bottomrule
\end{tabular}
\caption{Ranking variant ablation on GM-CIHT (\%). \textbf{$\Delta$} is the
change in micro-averaged F1 relative to the default ranking (cosine similarity
with proposition bias).}
\label{tab:ablation-ranking}
\end{table}

\paragraph{Entity detection: regex vs.\ NER.}
Entity spans in proposition extraction use the same lightweight pattern rules as Section~\ref{sec:entity-aware}. We replace them with a pre-trained NER system for boundaries, keeping the same ranking (cosine similarity with proposition bias). Table~\ref{tab:ablation-ner} reports GM-CIHT F1: NER reaches $81.5\%$, $-1.1$ below regex ($82.6\%$). GM-CIHT often contains explicit markers, capitalization, and hyphenation that regex handles reliably; NER can introduce false positives or boundary errors that hurt minimal-span proposition extraction. NER may still help on corpora with less regular surface forms; the accuracy-latency trade-off also matters for deployment.

\begin{table}[t]
\centering
\small
\begin{tabular}{lcc}
\toprule
\textbf{Entity detection} & \textbf{F1} & \textbf{$\Delta$} \\
\midrule
Regex (default) & 82.6 & --- \\
NER & 81.5 & $-$1.1 \\
\bottomrule
\end{tabular}
\caption{Entity detection variant on GM-CIHT (\%). \textbf{$\Delta$} is the
change in micro-averaged F1 relative to the default regex-based entity
detection.}
\label{tab:ablation-ner}
\end{table}

\subsection{Performance of Semantic Chunking in Different Settings}
\label{sec:analysis}

We analyze dataset-dependent performance patterns to understand when semantic
chunking is most effective.

\paragraph{Success on GM-CIHT and DDI.}
Both datasets contain diverse relation types with relatively clear semantic
boundaries (e.g., \textsc{treats} vs.\ \textsc{inhibits} vs.\ \textsc{coexists\_with}).
Tiered trigger prioritization effectively disambiguates these categories, while
entity-aware segmentation prevents fragmentation of complex biomedical terms.
The observed gains on GM-CIHT (+8.4 F1) and DDI (+1.0 F1) indicate that semantic
chunking is particularly beneficial for extraction tasks with explicit relation
signals and moderate entity spacing.

\paragraph{ChemProt: Entity Density Challenges.}
ChemProt exhibits high entity density, with multiple chemicals and proteins often
appearing within short spans (5--10 tokens). Entity-aware chunking can group
multiple entity pairs into a single chunk (e.g., \textit{``compound X inhibits
protein A and protein B''}), increasing ambiguity during relation assignment.
Moreover, ChemProt requires fine-grained biochemical distinctions
(e.g., \textsc{antagonist} vs.\ \textsc{downregulator}) that rely on broader
contextual cues beyond trigger words. Incremental ablations on ChemProt show an inverse pattern to GM-CIHT: adding tiered triggers and the tuned hierarchy can hurt F1, consistent with priorities designed for therapeutic relations misaligning with biochemical relation types. In this setting, fixed chunking may incidentally isolate simpler entity pairs, yielding slightly stronger performance.

\paragraph{ADE: Classification vs.\ Extraction.}
ADE is formulated as a binary classification task rather than structured
relation extraction. Performance therefore depends on broad contextual signals
such as symptom descriptions and temporal cues, rather than precise
subject--trigger--object spans. Proposition-focused chunking may omit such
supporting context, whereas fixed chunking preserves heterogeneous evidence
useful for holistic classification.

\paragraph{Reproducibility.}
Our BioMedRAG Fixed GM-CIHT F1 (74.2\%) is below the F1 reported in the original BioMedRAG publication (81.42\%), likely due to differences in hardware, hyperparameters (e.g., shorter maximum sequence length when training the chunk scorer under GPU memory limits), and our own pipeline integration and dataset format unification. The comparison between fixed and semantic chunking remains meaningful because both conditions use the same scorer, generator, and preprocessing.

\paragraph{Generalization.}
Overall, semantic chunking is most effective when (i) relation types are coarse-grained and trigger-explicit, (ii) entity density is moderate, and (iii) the task emphasizes structured extraction over classification. Because our evaluation intentionally follows the four BioMedRAG benchmarks, these conclusions are best interpreted for sentence-level biomedical relation extraction over published literature rather than for clinical notes or document-level retrieval. The configuration-driven design improves transparency and reuse, but it still requires task-specific symbolic resources, such as trigger lexicons and relation hierarchies, which may limit scalability when transferring to many unseen biomedical domains. High-density corpora and longer-context settings may benefit from future extensions incorporating adaptive entity-pair isolation, pair-specific proposition selection, automatic trigger induction, or variable chunk granularity.

\subsection{Error Analysis}
\label{sec:error-analysis}

We categorize the main failure modes of the proposed semantic chunking framework according to their likely root causes. First, errors occur when a relation is expressed without an explicit lexical trigger. In such cases, proposition-first extraction may not construct a complete subject--trigger--object span, and the framework relies on entity-aware or fixed fallback chunks. Second, high entity density can increase ambiguity, particularly in ChemProt, where multiple chemicals and proteins may appear within a short sentence. This makes it difficult to assign the correct relation to the correct entity pair. Third, fine-grained biochemical relations may require broader mechanistic context than a minimal proposition span provides. Relations such as \textsc{Agonist}, \textsc{Activator}, and \textsc{Downregulator} can share similar surface cues while differing in biological meaning. Finally, ADE differs from the extraction datasets because binary adverse-event classification often depends on sentence-level context, including symptoms, temporal expressions, speculation, and negation. These failure modes explain why semantic chunking improves datasets with explicit relation cues and moderate entity density, while fixed-size chunking remains competitive for dense extraction and classification settings.

\section{Conclusion}
\label{sec:conclusion}

Fixed-size chunking in retrieval-augmented biomedical relation extraction
frequently fragments entities and splits relation-bearing propositions,
degrading the quality of retrieved evidence for downstream generation.

We introduced a configurable semantic chunking framework that combines
entity-aware sliding windows with proposition-first extraction.
Tiered trigger prioritization organizes relation signals by semantic
specificity, while relation hierarchy resolution disambiguates competing
interpretations by encoding task-specific priorities. All components are
externalized in configuration files, enabling adaptation to new biomedical
domains without code modification.

Our approach yields substantial gains on extraction-focused benchmarks,
achieving +8.4 F1 points on GM-CIHT (82.6\%) and +1.0 F1 on DDI (79.2\%).
Ablation studies show that relation hierarchy resolution contributes the
largest incremental gain in the GM-CIHT stack (+3.9 F1), and that semantically coherent chunks peak with a single in-context example (Table~\ref{tab:ablation-icl}). The ranking ablation (Table~\ref{tab:ablation-ranking}) shows that proposition bias outperforms mixing structural scores into cosine ranking; the NER ablation (Table~\ref{tab:ablation-ner}) shows that pattern-based entity spans match or beat NER on GM-CIHT under our setup.

Absolute F1 for the fixed baseline is below published BioMedRAG figures, but the fixed-versus-semantic comparison is conducted under identical training and preprocessing and remains the primary evidence for our claims.

Analysis reveals that semantic chunking is most effective for datasets with
coarse-grained relations and moderate entity density. In contrast, for
high-density corpora (ChemProt) or classification-oriented tasks (ADE),
fixed-size chunking remains competitive, as entity preservation may group
multiple targets and proposition extraction can over-constrain contextual
evidence.

Future work includes adaptive chunking strategies that respond to entity
density, dataset-specific hierarchy design (or disabling hierarchy) for
fine-grained biochemistry tasks, extension to clinical notes and drug
labels, and entity-pair-specific isolation mechanisms to better handle
multi-target scenarios. Future releases of the integration code,
configuration files, and unified data conversion scripts will further
support reproducibility and facilitate comparison with future
BioMedRAG-based systems.

\section*{Code and Data Availability}

All datasets used in this study are publicly available through the BioMedRAG benchmark setting and the original dataset sources. To support reproducibility, we will release the semantic chunking implementation, dataset configuration files, preprocessing scripts, unified JSONL conversion format, and evaluation instructions upon publication. The configuration files include trigger vocabularies, tier assignments, relation hierarchies, negation rules, context markers, and chunking parameters. These resources are intended to reproduce the fixed-size and semantic chunking comparisons reported in this work under the same BioMedRAG scorer and generator pipeline.

\section*{Acknowledgements}
Funded by the Deutsche Forschungsgemeinschaft (DFG, German Research
Foundation) -- 527049502.

\bibliographystyle{named}
\bibliography{References}

\end{document}